\documentclass{article}
\usepackage{spconf,amsmath,graphicx}
\makeatletter
\renewenvironment{abstract}{%
  \if@twocolumn
    \begin{center}\bfseries ABSTRACT\end{center}%
  \else
    \begin{center}\bfseries ABSTRACT\end{center}%
  \fi
  \small\noindent
}{\par}
\makeatother
\makeatother
\usepackage{amssymb} % for \varnothing
\usepackage{tikz} % Required for \usetikzlibrary
\usetikzlibrary{arrows.meta,positioning}
\usetikzlibrary{arrows.meta,positioning,calc}
\usepackage{bbm}  % in preamble
\usepackage[bb=boondox]{mathalfa}
\usepackage{cite}
\usepackage{graphicx} % in preamble
\usepackage{booktabs} % for \toprule, \midrule, \bottomrule
\usepackage{multirow} % for \multirow in tables
\usepackage{graphicx}
\usepackage{siunitx} % for S columns in tables
\usepackage{float} % allows [H] float placement
\usepackage{dblfloatfix}  % fixes ordering + improves dbl float top placement
\usepackage{booktabs}      % for \toprule, \midrule, \cmidrule, \bottomrule
\usepackage{siunitx}       % for S columns (already used in your table)
\usepackage{caption}
\usepackage[section]{placeins} % \FloatBarrier works per-section
\usepackage{xcolor}
\usepackage{tikz}
\usetikzlibrary{arrows.meta,positioning,calc,shapes.geometric}
\usepackage{placeins}     % for \FloatBarrier
\usepackage{caption}
\usepackage{dblfloatfix}   % fixes some edge cases for figure*/table*
\makeatletter
\def\fps@figure{t}  % figures prefer top
\def\fps@table{t}   % tables prefer top
\makeatother

\usepackage{dblfloatfix}  % fixes some 2-col float edge 

\title{SARNet: A Spike-Aware consecutive validation Framework for Accurate Remaining Useful Life Prediction}
\name{
  Junhao Fan$^{1\dagger}$,
  Wenrui Liang$^{2\dagger}$,
  Wei-Qiang Zhang$^{2*}$
  \thanks{$\dagger$ Equal contribution. *Corresponding author.}
  \thanks{This work was supported by the National Natural Science Foundation of China under Grant No. 62276153.}
}
\address{
$^{1}$Georgetown University, Washington, DC, USA \\
$^{2}$Department of Electronic Engineering, Tsinghua University, Beijing, China \\
{\tt\small jf1687@georgetown.edu, wqzhang@tsinghua.edu.cn}
}
\begin{document}
\ninept
\maketitle

\begin{abstract}
Accurate prediction of remaining useful life (RUL) is essential to enhance system reliability and reduce maintenance risk. Yet many strong contemporary models are fragile around fault onset and opaque to engineers: short, high-energy spikes are smoothed away or misread, fixed thresholds blunt sensitivity, and physics-based explanations are scarce. To remedy this, we introduce \textbf{SARNet} (Spike-Aware Consecutive Validation Framework), which builds on a Modern Temporal Convolutional Network (ModernTCN) and adds spike-aware detection to provide physics-informed interpretability. ModernTCN forecasts degradation-sensitive indicators; an adaptive consecutive threshold validates true spikes while suppressing noise. Failure-prone segments then receive targeted feature engineering (spectral slopes, statistical derivatives, energy ratios), and the final RUL is produced by a stacked RF--LGBM regressor. Across benchmark-ported datasets under an event-triggered protocol, SARNet consistently lowers error compared to recent baselines (RMSE 0.0365, MAE 0.0204) while remaining lightweight, robust, and easy to deploy.

\end{abstract}

\begin{keywords}
Remaining useful life (RUL), predictive maintenance, modern temporal convolutional networks (MTCN), spike-aware onset detection
\end{keywords}

\section{Introduction}
Remaining useful life (RUL) prediction is central to predictive maintenance, especially for bearings whose degradation can accelerate after onset.~\cite{Ferreira2022RULReview, article, Magadan2024RULDL} Classical statistical thresholds rarely cope with the non-linear, condition-dependent dynamics seen in practice ~\cite{IE1, IE2, IE3}.

Recent research spans both event detection and deep sequence modeling. Chen and Tian~\cite{Chen_and_tian} use a prototypical network to detect critical points (sudden degradation) and then switch to RUL prediction, updating the reference state over time. Yao et al.~\cite{Yao} build Patch ModernTCN-Mixer, a dual-task network that couples first prediction time (FPT) detection with RUL using a dynamic semi-soft threshold and GradNorm. Other pipelines pair signal denoising and hand-crafted health indices with recurrent predictors. For example, Zou et al.~\cite{Zou2025ADBilstmRUL} denoise with DWT, rank the candidate features, compress them with KPCA, and then feed the resulting embeddings to an attention-DBiLSTM, achieving strong cross-dataset robustness. Convolutional approaches further improve accuracy~\cite{Bai2018Empirical, Yang2024CNNVAE_MBiLSTM, Qiu2023PiecewiseTCN}. A TCN with causal dilations plus multi-head self-attention reweights the global context for rolling-bearing RUL~\cite{Jiang2024TCNMSA, Zhang2024ResidualTCNAttentionRUL}; end-to-end multi-channel CNNs, optionally followed by attention LSTM, deliver strong MAE/RMSE from raw signals~\cite{Jiang2020TSMC_CNN_ALSTM}.

However, three practical gaps persist. First, fixed spike rules (e.g., $x{+}3\sigma$) are noise-sensitive and often trigger prematurely~\cite{ZhangWu2005SuppRunsRules}. Second, even strong deep features usually require calibration to yield stable post-onset RUL when degradation evidence is sparse ~\cite{Biggio2021UncertaintyAwareRUL}. Third, results on bearing life prediction should have the opportunity to be applied in real-life conditions. However, although most deep learning models and other models have accurate prediction results, they lack model interpretability and are difficult to deploy in real life.~\cite{c3_1, Serradilla2020PredictiveMaintenanceSurvey, Fink2020PHMDeepLearning}

\textbf{SARNet} addresses both issues with an event-triggered, two-stage design. A Modern Pure Convolution Structure for General Time Series (MTCN)~\cite{MTCN} first forecasts a degradation-sensitive indicator; an \emph{adaptive consecutive-spike} rule then confirms sustained onset before prognosis. Unlike prior first-prediction time (FPT) detectors by Yao et al., our rule demands $d_{\min}$ consecutive exceedances and \emph{falls back} to full-sequence prediction when spike evidence is weak, reducing false alarms without sacrificing coverage. For the post-onset segment, we replace a single heavy predictor with a lightweight, interpretable RF-LGBM ensemble head, providing calibrated estimates and feature attributions. By separating forecasting, onset validation, and calibrated regression, we concentrate learning where it matters most, reduce sensitivity to noise, and keep the whole pipeline interpretable.

\paragraph*{Contributions are listed below:}
\begin{itemize}
  \item\textbf{Event-triggered formulation.} Rather than one–shot RUL regression, ModernTCN forecasts a degradation indicator \(\hat z_{t+h}=f_\theta(x_{t-L+1:t})\). A spike-aware detector declares onset only if
\[
T_t=\bigwedge_{i=0}^{K-1}\bigl(\hat z_{t-i} > \tau_t\bigr),\quad 
\tau_t=\mu_t + k_\sigma\,\sigma_t ,
\]
with a small hysteresis to avoid chattering. This \emph{consecutive, adaptive} rule encodes impact-induced transients observed in vibration signals, tying triggers to fault physics and making the subsequent RUL estimation interpretable and robust across conditions.
  \item \textbf{Robust onset validation.} An adaptive consecutive–spike rule with a simple fallback (when evidence is weak) confirms onset, making the trigger far less noisy than fixed \(x+3\sigma\) thresholds.
  \item \textbf{Calibrated and interpretable prognosis.} After onset is validated, a lightweight, interpretable RF-LGBM head delivers calibrated post–onset RUL with clear importance of characteristics, facilitating real engineering use.
\end{itemize}

\section{Methodology}
\label{sec:format}

This study proposes the SARNet to predict the remaining useful life (RUL). The framework integrates a Modern Temporal Convolutional Network (ModernTCN), an adaptive spike detection mechanism, and ensemble regression models. Figure 1 provides an overview of the pipeline.

\begin{figure}[t]
  \centering
  \includegraphics[width=1.01\columnwidth]{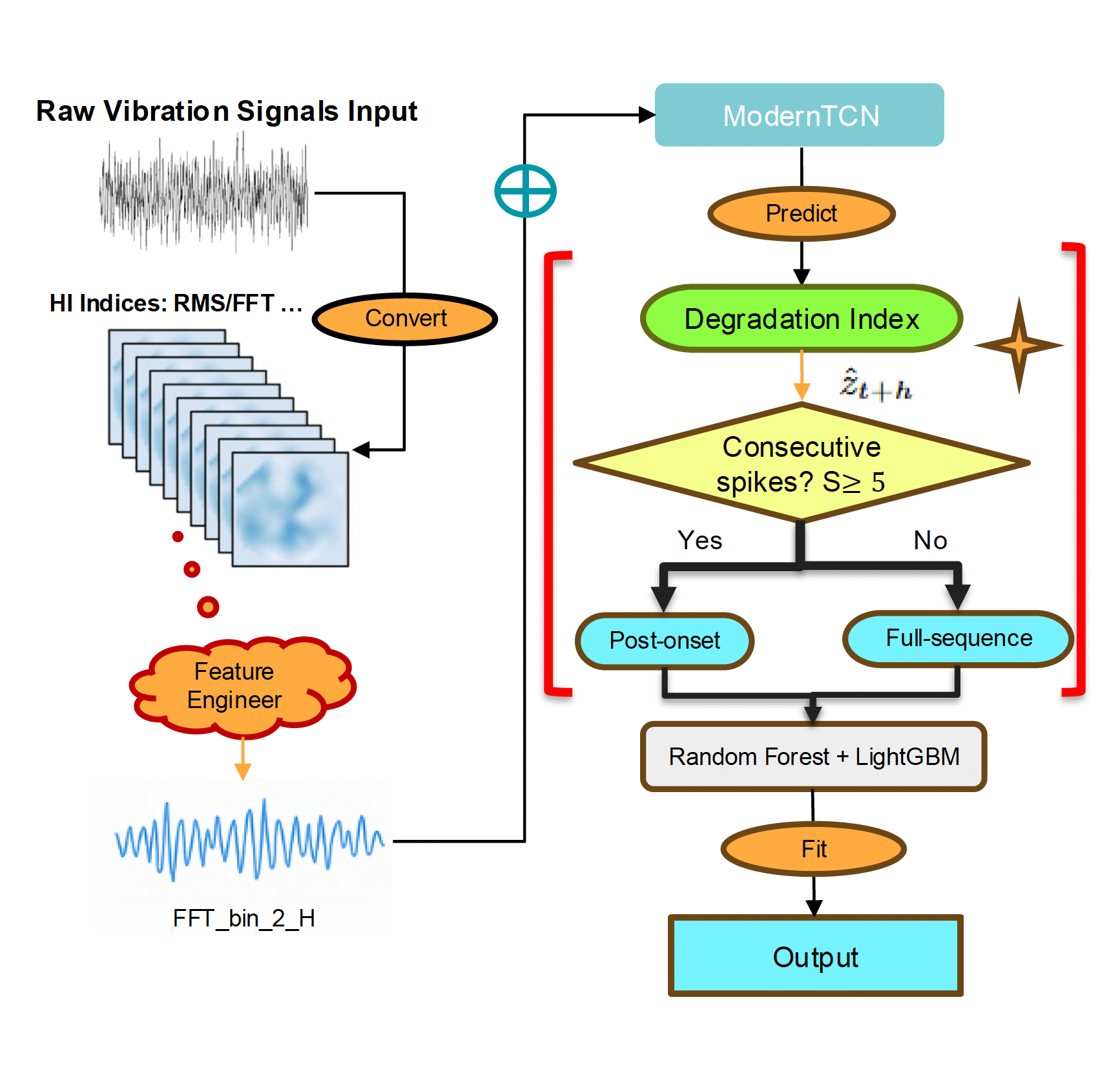}
  \caption{The SARNet Framework}
  \label{fig:arch}
\end{figure}

% ---------------- Single-column version (recommended) -------

The raw sensor signals are denoted as $x_t \in \mathbb{R}$, where $t$ indexes discrete time steps with a sampling interval of $\Delta t = 60$ seconds. 

To identify the most degradation-sensitive feature, we computed the absolute Spearman correlation between each engineered feature and the ground-truth RUL under all operating conditions. As shown in Table 1, \texttt{FFT\_bin\_2\_H}, which is denoted as the value extracted from the second frequency bin of the magnitude of FFT of the horizontal signal, achieved the highest absolute correlation, making it the main indicator used for subsequent modeling.

\begin{table}[h]
\centering
\begin{tabular}{l c}
\hline
\textbf{Feature Name} & \textbf{Absolute Spearman Correlation} \\
\hline
FFT\_bin\_2\_H & 0.535 \\
Skewness\_H & 0.302 \\
FFT\_bin\_5\_H & 0.301 \\
BandPower\_Mid\_V & 0.272 \\
Kurtosis\_V & 0.242 \\
BandPower\_Mid\_H & 0.203 \\
RMS\_V & 0.202 \\
Kurtosis\_H & 0.182 \\
FFT\_bin\_4\_V & 0.179 \\
CrestFactor\_H & 0.168 \\
\hline
\end{tabular}
\caption{Top 10 sensor features ranked by absolute Spearman correlation with RUL.}
\label{tab:feature_correlation}
\end{table}

In this research, each feature is normalized using min-max scaling:
\begin{equation}
x_t^{\text{norm}} = \frac{x_t - \min(x)}{\max(x) - \min(x)},
\end{equation}
where $\min(x)$ and $\max(x)$ are calculated in the training set to avoid data leakage. The data set is then divided into training and testing partitions, ensuring that there is no temporal overlap between them.

\subsection{Adaptive Consecutive Spike Detection}
Accurate detection of degradation onset is critical for reliable RUL prediction, particularly under varying operating conditions where transient noise may cause false alarms. Conventional thresholding methods, such as the fixed $x + 3\sigma$ rule \cite{Lehmann2013ThreeSigma}, treat any single exceedance as an anomaly, making them highly sensitive to random fluctuations.

Our new model, SARNet, overcomes this limitation and provides mechanism-level interpretability consistent with engineering and physical principles.

\subsubsection{Adaptive Thresholding}
Given a predicted degradation-sensitive feature sequence $\{\hat{y}_t\}_{t=1}^T$, the adaptive threshold $\theta$ is defined as:
\begin{equation}
\theta = \mu_{\text{ref}} + k\sigma_{\text{ref}}, \quad k=2
\end{equation}
where $\mu_{\text{ref}}$ and $\sigma_{\text{ref}}$ are the mean and standard deviation computed from a reference (healthy) operating window, and $k$ is a sensitivity coefficient. Compared to the traditional $k=3$ setting, $k=2$ allows earlier detection while relying on subsequent validation to avoid false positives.

\subsubsection{Subsequent Consecutive Spike Validation}
A spike event is only confirmed if the feature value exceeds $\theta$ for at least $d_{\min}$ consecutive time steps:
\begin{equation}
\sum_{j=0}^{d_{\min}-1} \mathbb{I}\left[ \hat{y}_{t+j} > \theta \right] = d_{\min}
\end{equation}
where $\mathbb{I}[\cdot]$ is the indicator function. This sustained-exceedance criterion filters out short-lived fluctuations caused by noise or transient load changes.

\subsubsection{Fallback Mechanism}
If the total number of validated spikes $N_{\text{spike}}$ is below a minimum threshold $n_{\min}$ (empirically set to 5), the system returns to using the full ModernTCN prediction sequence for the estimation of RUL:
\begin{equation}
\text{RUL Input} =
\left\{
\begin{aligned}
&\text{Features within spike window}, && N_{\text{spike}} \geq n_{\min} \\
&\text{Entire predicted sequence}, && N_{\text{spike}} < n_{\min}
\end{aligned}
\right.
\label{eq:fallback}
\end{equation}
This ensures stable performance in cases with weak or delayed degradation signatures.

\subsubsection{Advantages}
The proposed adaptive thresholding with consecutive spike validation and fallback ensures robustness by reducing noise-induced false alarms, retaining only sustained degradation trends for RUL regression, or reverting to the full predicted sequence when spike evidence is insufficient.

This design improves both robustness and timeliness in detecting the onset of degradation, directly benefiting the final accuracy of the RUL regression. Consistent with the run-rules theory in Statistical Process Control (SPC) \cite{Oh2020IndividualsRunsRules}, we require at least \(d_{\min}=5\) consecutive spikes to validate the onset, which reduces noise-driven false alarms and preserves sensitivity to sustained degradation; otherwise, we return to full-sequence prediction.

\begin{equation}
\mathcal{S} = \{ t \mid \hat{x}_t > \tau \}.
\end{equation}

To be more specific, if the number of detected spikes is $|\mathcal{S}| \geq 5$, only these spike segments are retained for subsequent processing. If $|\mathcal{S}| < 5$, the algorithm comes back to using the entire predicted sequence $\{\hat{x}_t\}_{t=t_0}^{T_f}$ to ensure adequate data coverage. 

This design reduces false positives from transient noise and ensures that the identified onset corresponds to a sustained degradation trend rather than a short-lived fluctuation. It also improves model robustness under cross-condition scenarios, where noise profiles can differ significantly.

\subsection{Post-Onset RUL Estimation Pipeline}
After detecting the onset of degradation $t_s$ via the adaptive consecutive spike detection module, only the post-onset segment is used for RUL regression. The normalized RUL at time $t \ge t_s$ is defined as:
\begin{equation}
\mathrm{RUL}_{\text{seg}}(t) = \frac{T_f - t}{T_f - t_s},
\end{equation}
where $T_f$ is the failure time, ensuring $\mathrm{RUL}_{\text{seg}}(t_s) = 1$ and $\mathrm{RUL}_{\text{seg}}(T_f) = 0$. For each $t$, the feature vector $\mathbf{f}_t$ includes the indicator predicted by ModernTCN ($\hat{x}_t$), the spectral slopes and peak magnitudes, the statistical derivatives (e.g., \texttt{FFT\_bin\_2\_H} slope, variance, energy) and the time-domain spike metrics. These features are input to a stacking ensemble combining Random Forest (RF) and LightGBM (LGBM), with a Ridge meta-learner weighting their outputs:
\begin{equation}
\widehat{\mathrm{RUL}}_{\text{seg}}(t) = \alpha \cdot \widehat{\mathrm{RUL}}_{\text{RF}}(t) + (1 - \alpha) \cdot \widehat{\mathrm{RUL}}_{\text{LGBM}}(t),
\end{equation}
where $\alpha$ is learned from validation data. This integration takes advantage of the robustness of the RF and the efficiency of LGBM to improve the estimation of RUL.

This integration takes advantage of the robustness of RF and the efficiency of LGBM to improve RUL estimation, particularly in the post-onset prediction scenario, where precise modeling of degradation trends is critical.

\subsection{Evaluation metrics}\label{sec:metrics}
We report the mean absolute error (MAE), the root mean square error (RMSE), the coefficient of determination ($R^2$), and the PHM~2012 scoring function. Metrics are computed on normalized RUL series $\{y_t\}_{t=1}^m$ with predictions $\{\hat y_t\}_{t=1}^m$ and $\bar y=\frac{1}{m}\sum_{t=1}^m y_t$:
\begin{align}
\mathrm{MAE} &= \frac{1}{m}\sum_{t=1}^{m}\left|y_t-\hat y_t\right|, \label{eq:mae}\\
\mathrm{RMSE} &= \sqrt{\frac{1}{m}\sum_{t=1}^{m}\left(y_t-\hat y_t\right)^2}, \label{eq:rmse}\\
R^2 &= 1-\frac{\sum_{t=1}^{m}\left(\hat y_t-y_t\right)^2}{\sum_{t=1}^{m}\left(\bar y-y_t\right)^2}. \label{eq:r2}
\end{align}

Following the IEEE PHM~2012 Data Challenge~\cite{Nectoux2012PRONOSTIA}, the Score penalizes late and early predictions asymmetrically:
\begin{align}
\mathrm{Score} &= \frac{1}{n-1}\sum_{t=1}^{n-1} A_t, \label{eq:score}\\
A_t &= 
\begin{cases}
\exp\!\Big(-\ln(0.5)\,\dfrac{Er_t}{5}\Big), & Er_t \le 0,\\[6pt]
\exp\!\Big(+\ln(0.5)\,\dfrac{Er_t}{20}\Big), & Er_t > 0,
\end{cases} \label{eq:At}\\
Er_t &= \dfrac{y_t-\hat y_t}{y_t}\times 100\% . \label{eq:er}
\end{align}

For completeness, a common construction of the ground-truth RUL $y_t$ over a run of length $T$ with first-prediction time $\mathrm{FPT}$ is
\begin{equation}
y_t =
\begin{cases}
T, & 0\le t \le \mathrm{FPT},\\[2pt]
T - T\,\dfrac{t-\mathrm{FPT}}{T-\mathrm{FPT}}, & \mathrm{FPT}\le t \le T.
\end{cases}
\end{equation}

\section{Experiments}
\label{sec:pagestyle}
\subsection{Dataset}
We evaluate our framework on the XJTU\textendash SY bearing accelerated life dataset~\cite{WangHybrid}, which provides run\textendash to\textendash failure vibration data for 15 bearings collected under three operating conditions. Each record is acquired by two orthogonal accelerometers (horizontal and vertical) at a sampling rate of 25.6\,kHz, with 32\,768 points per channel captured every 1\,min, yielding long sequences that cover the full degradation process. For this study, we focus on the horizontal channel and use the engineered spectral indicator \texttt{FFT\_bin\_2\_H}, which our Spearman analysis identified as the feature most sensitive to RUL among the available candidates.

\subsection{Experimental Setup}
Following the methodology described in Sec.~2, we conduct experiments on the XJTU-SY bearing dataset using a single NVIDIA GeForce RTX 3090 GPU with 24\,GB memory and 18 CPU cores. The missing values in \texttt{FFT\_bin\_2\_H} are removed, and the remaining useful life (RUL) is calculated as a backward count from the failure time for each bearing run.  

A Modern Temporal Convolutional Network with three residual blocks ([32, 32, 16] channels) and exponentially increasing dilations ($2^i$) is trained in the \texttt{FFT\_bin\_2\_H} feature using a sequence length $SEQ\_LEN = 20$ and a prediction horizon $PRED\_STEP = 5$ minutes. The model is optimized with Adam ($\eta = 10^{-3}$) and MSE loss for 10 epochs, with a batch size of 32. 

\section{Results}

% ---------- Table 1: Performance (single-column) ----------
\begin{table}[t]
\centering
\caption{Performance under segmentation and full-length evaluation for all test bearings.}
\label{tab:perf}
\setlength{\tabcolsep}{9pt}
\resizebox{\columnwidth}{!}{%
\begin{tabular}{cccccccccccc}
\toprule
\multirow{2}{*}{Test Bearing} & \multirow{2}{*}{$k_\sigma$} &
\multicolumn{4}{c}{Segmentation Mode} & &
\multicolumn{4}{c}{Full-length Mode} \\
\cmidrule(l){3-6}\cmidrule(l){8-11}
& & RMSE & MAE & $R^2$ & MAPE & & RMSE & MAE & $R^2$ & MAPE \\
\midrule
bearing1\_3 & 2 & 0.0308 & 0.0196 & 0.9888 & 661.30 && 0.0486 & 0.0291 & 0.9720 & 480.03 \\
bearing1\_3 & 3 & 0.0400 & 0.0232 & 0.9811 & 405.65 && 0.0587 & 0.0349 & 0.9591 & 296.95 \\
bearing2\_5 & 2 & 0.0337 & 0.0190 & 0.9865 & 1377.31 && 0.0692 & 0.0329 & 0.9429 & 1192.20 \\
bearing2\_5 & 3 & 0.0367 & 0.0207 & 0.9839 & 1121.56 && 0.0687 & 0.0339 & 0.9438 & 971.15 \\
bearing3\_5 & 2 & 0.0754 & 0.0564 & 0.9332 & 8236.07 && 0.1049 & 0.0747 & 0.8702 & 5129.44 \\
bearing3\_5 & 3 & 0.0678 & 0.0513 & 0.9461 & 6615.39 && 0.0972 & 0.0686 & 0.8886 & 4120.97 \\
\bottomrule
\end{tabular}}
\end{table}

% ---------- Table 2: Ablation (single-column) ----------
\begin{table}[t]
\centering
\setlength{\tabcolsep}{9pt}
\caption{Ablation of the proposed \textbf{SAR} (Spike-Aware) framework. Spike-aware variants (top) consistently improve $R^2$ / RMSE despite lower coverage (fewer, high-signal windows). Best model in bold.}
\label{tab:esar-ablation}
\resizebox{\columnwidth}{!}{%
\begin{tabular}{
  l
  S[table-format=1.6]
  S[table-format=1.6]
  S[table-format=1.6]
  r
  l
  r
}
\toprule
\textbf{Method} & \textbf{RMSE} & \textbf{MAE} & $\boldsymbol{R^2}$ & \textbf{Coverage} & \textbf{Learner} & \textbf{\# of Feature} \\
\midrule
\multicolumn{7}{l}{\textit{SAR Framework (Spike-Aware Detection + MTCN backbone)}}\\
\textbf{SARNet}          & {\bfseries 0.036187} & {\bfseries 0.020585} & {\bfseries 0.989434} & 118  & ENS    & 9 \\
SAR + RF          & 0.037329 & 0.014730 & 0.988756          & 118  & RF     & 9 \\
SAR + LGBM        & 0.044625 & 0.029210 & 0.983931          & 118  & LGBM   & 9 \\
SAR (linear head) & 0.312706 & 0.263490 & 0.210959          & 118  & Linear & 1 \\
\addlinespace
\multicolumn{7}{l}{\textit{Ablations without Spike-Aware Detection (MTCN backbone only)}}\\
MTCN + LGBM         & 0.067023 & 0.043393 & 0.946094          & 1833 & LGBM   & 9 \\
MTCN + RF--LGBM     & 0.075449 & 0.046989 & 0.931689          & 1833 & ENS    & 9 \\
MTCN + RF           & 0.088960 & 0.054662 & 0.905033          & 1833 & RF     & 9 \\
MTCN (linear head)  & 0.288225 & 0.251415 & 0.003116          & 1833 & Linear & 1 \\
\bottomrule
\end{tabular}}
\end{table}% keep all three tables inside the Results section

\newcommand{\COneADRMSE}{0.0682}
\newcommand{\COneADMAE}{0.0741}
\newcommand{\COneADRtwo}{0.8743}

\newcommand{\CTwoADRMSE}{0.0811}
\newcommand{\CTwoADMAE}{0.0591}
\newcommand{\CTwoADRtwo}{0.9211}

\newcommand{\CThreeADRMSE}{0.0866}
\newcommand{\CThreeADMAE}{0.0533}
\newcommand{\CThreeADRtwo}{0.9100}
\subsection{Condition-Wise Evaluation with Sigma Threshold Comparison}
To make a fair, like-to-like comparison with the attention-DBiLSTM (A-DBiLSTM) study by Zou et al.~\cite{Zou2025ADBilstmRUL}, we keep the original condition-wise train/test splits on XJTU-SY and evaluate post-onset segmentation and full-length modes (Table~\ref{tab:perf}). Think of the protocol as a controlled trial: we ask both methods to run the same course, then watch what happens when faults begin. Across the three operating conditions, SARNet enters the post-onset window earlier and steadier, turning energy spikes into actionable triggers rather than noise; the result is a materially lower error and higher \(R^2\) than A-DBiLSTM without sacrificing whole-sequence reporting.
Concretely, under the same splits, SARNet's \emph{segmentation} mode cuts error markedly relative to A-DBiLSTM: RMSE drops by about 55\% in Condition~1, 58\% in Condition~2, and 22\% in Condition~3. Averaged across the three conditions, this is a 45\% reduction in RMSE; MAE follows the same pattern with reductions of roughly 74\%, 68\%, and a modest 4\% in the hardest setting. The full-length mode remains favorable in two conditions and within range in the third, but the gain primarily comes from spike-aware segmentation. The takeaway is straightforward: Once the onset is validated, spike-aware selection reduces variance, dampens false alarms, and stabilizes predictions, giving practitioners a clearer picture of remaining life while preserving traceability to the raw signal. And avoids overfitting to condition shifts seen in practice.

The other takeaway is also intuitive. Using $\sigma=2$ with SAR framework produces markedly lower errors than $\sigma=3$, substantially improving the predictive capacity of the model. This confirms that our refinement of the $3\sigma$ rule is reasonable and effective. In practice, the tighter threshold captures incipient spikes earlier and reduces variance in the post onset window. It also reduces false alarms while keeping the predictions traceable to the original signal.
% =====================================================The takeaway is simple: Once the onset is validated, spike-aware selection reduces variance, dampens false alarms, and stabilizes predictions, giving practitioners a clearer picture of remaining life while preserving traceability to the raw signal. And avoids overfitting to condition shifts seen in practice.

\subsection{Ablation Study}
Table~\ref{tab:esar-ablation} presents the ablation study. The spike aware selection combined with the ModernTCN backbone produces consistent gains regardless of the regression head. The best variant, \textbf{SARNet}, which averages the predictions of Random Forest and LightGBM, achieves a 0.036 RMSE, 0.021 MAE, and approximately 99 percent \(R^2\). Using a single head remains competitive, but is weaker: SAR with Random Forest achieves RMSE $0.037329$ and $R^2$ $0.988756$, and SAR with LightGBM reaches RMSE $0.044625$ and $R^2$ $0.983931$. A linear head under the same backbone performs poorly, which underlines the need for a nonlinear regressor in the post onset regime.

When spike aware detection is removed, the accuracy drops markedly even though many more windows are used. The strongest non–spike baseline (MTCN with LightGBM) yields RMSE $0.067$ and $R^2$ $0.946$. Compared to this baseline, SARNet reduces RMSE by approximately \textbf{46\%} and increases $R^2$ by approximately \textbf{0.043}. Notably, SAR variants operate on \emph{118} post-onset windows on average, whereas the full length mode uses \emph{1833} windows. This corresponds to about fifteen times fewer inputs while delivering better accuracy, which is attractive for real time use.

\subsection{Comparison with Previous Researches}
Figure~\ref{fig:arch} compares the SARNet with representative predictors of RUL bearings from recent years. Under the same experimental protocol and evaluation metrics (RMSE, MAE, MAPE, and \(R^{2}\)), our approach remains lightweight while achieving better performance, with consistently higher \(R^{2}\) and lower RMSE, MAE, and MAPE than competing methods.

\begin{figure}[t]
  \centering
  \includegraphics[width=1.01\columnwidth]{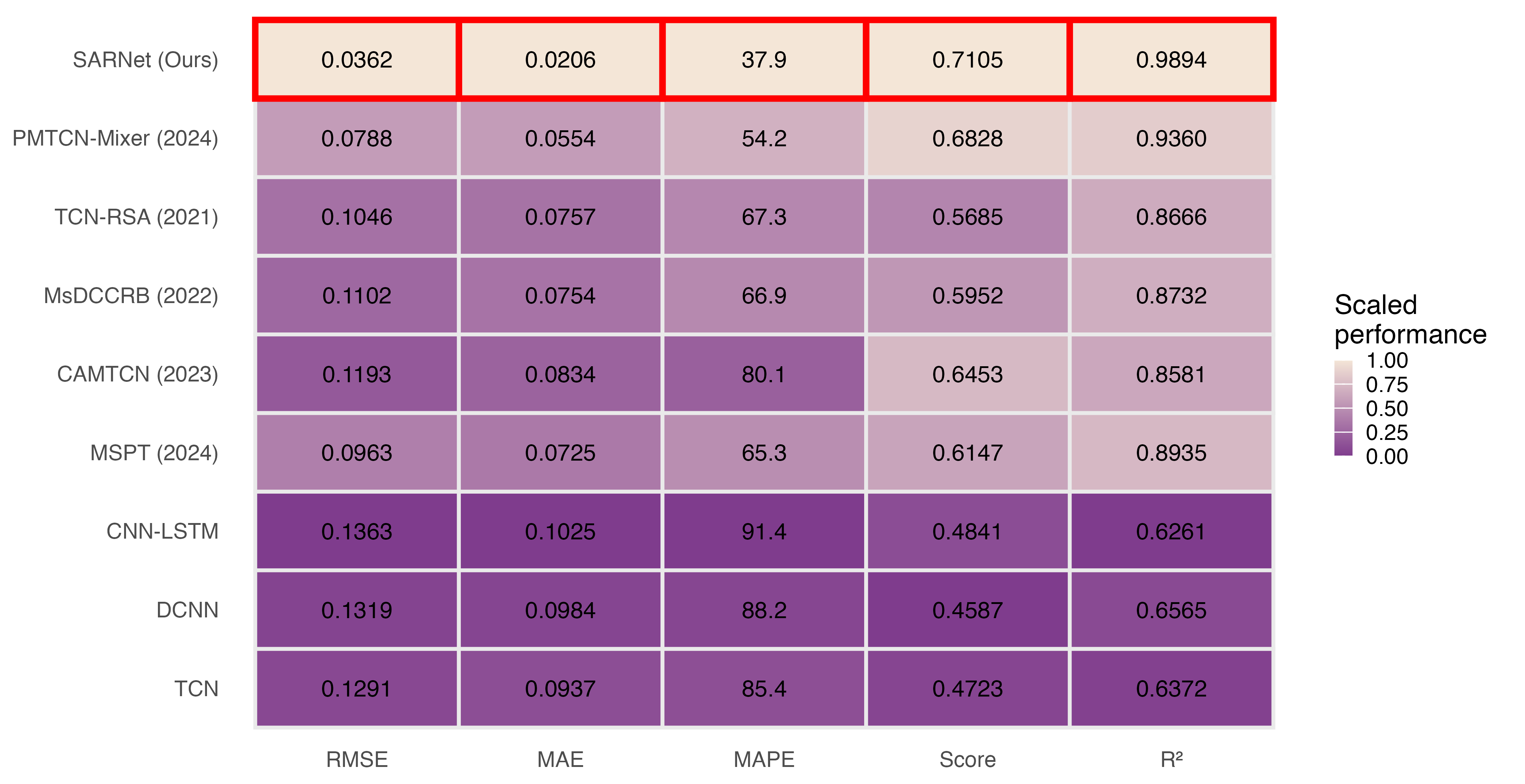}
  \caption{Comparison with representative baselines (baselines numbers from \cite{Yao}); our results follow the same protocol.}
  \label{fig:arch}
\end{figure}

In general, three conclusions emerge. 
\begin{itemize}
  \item The \textbf{Post onset evaluation is preferable,} as the concentration of the assessment on post onset segments consistently outperforms the full sequence scoring across bearings and metrics.
  \item \textbf{Spike-Aware detection is the main driver,} which yields substantial gains even before the downstream regressor is chosen. 
  \item The \textbf{RF\textsc{-}LGBM Ensemble is the most reliable combination} within the SAR family, combining Random Forest and LightGBM produces the most stable and accurate post–onset RUL estimates, while preserving interpretability and ease of deployment.
\end{itemize}

\section{Conclusion}
\label{sec:print}
 SARNet demonstrates reliable remaining useful life prediction across the XJTU-SYbearings with consistent gains in accuracy and stability. The framework couples a ModernTCN forecaster with an adaptive consecutive spike validator, which suppresses noise while preserving abrupt degradation cues that are often diluted by end to end sequence models. Because the validator adapts to each run and condition, the method remains effective without requiring matched data across operating regimes and is therefore well suited to resource constrained deployments. 

The proposed model outperforms recent convolutional and recurrent baselines. In segmentation mode, it yields roughly a one half reduction in RMSE and a sixty percent reduction in MAE relative to a strong ModernTCN Mixer baseline, while achieving an $R^{2}$ close to $0.99$. These improvements arrive with clear feature attributions from the RF and LightGBM head and a small computational footprint that runs comfortably on a CPU. This progress not only lowers the deployment cost for enterprises, it also makes the model’s predictive mechanism easier to understand.

Overall, SARNet offers a practical, lightweight, and transparent alternative to black box deep networks for industrial prognostics, maintaining the degradation characteristics needed for precise post onset RUL estimation while remaining easy to interpret and deploy.

\vfill\pagebreak

% -------------------------------------------------------------------------
\bibliographystyle{IEEEbib}
\bibliography{strings,refs}

\end{document}